\def\year{2021}\relax
\documentclass[letterpaper]{article} 
\usepackage{aaai21}  
\usepackage{times}  
\usepackage{helvet} 
\usepackage{courier}  
\usepackage[hyphens]{url}  
\usepackage{graphicx} 
\usepackage{graphicx}  
\usepackage{natbib}  
\usepackage{caption} 
\usepackage{subcaption}
\usepackage{amsmath}
\usepackage{amssymb}
\usepackage{multirow}
\usepackage{tabularx}
\usepackage{booktabs}
\usepackage{xcolor}
\usepackage[switch]{lineno}

\newcolumntype{L}{>{$}l<{$}}
\newcolumntype{C}{>{$}c<{$}}
\newcolumntype{R}{>{$}r<{$}}

\title{IDOL: Inertial Deep Orientation-Estimation and Localization}
\author {
    Scott Sun,
    Dennis Melamed,
    Kris Kitani \\
}
\affiliations {
    Carnegie Mellon University
}

\begin{document}

\defcitealias{denoise_imu}{Brossard et.\ al. (2020)}

\maketitle

\begin{abstract}
Many smartphone applications use inertial measurement units (IMUs) to sense movement, but the use of these sensors for pedestrian localization can be challenging due to their noise characteristics. Recent data-driven inertial odometry approaches have demonstrated the increasing feasibility of inertial navigation. However, they still rely upon conventional smartphone orientation estimates that they assume to be accurate, while in fact these orientation estimates can be a significant source of error. To address the problem of inaccurate orientation estimates, we present a two-stage, data-driven pipeline using a commodity smartphone that first estimates device orientations and then estimates device position. The orientation module relies on a recurrent neural network and Extended Kalman Filter to obtain orientation estimates that are used to then rotate raw IMU measurements into the appropriate reference frame. The position module then passes those measurements through another recurrent network architecture to perform localization. Our proposed method outperforms state-of-the-art methods in both orientation and position error on a large dataset we constructed that contains 20 hours of pedestrian motion across 3 buildings and 15 subjects. Code and data are available at https://github.com/KlabCMU/IDOL.
\end{abstract}

\section{Introduction}
Inertial localization techniques typically estimate 3D motion from inertial measurement unit (IMU) samples of linear acceleration (accelerometer), angular velocity (gyroscope), and magnetic flux density (magnetometer). A well-known weakness that has plagued inertial localization is the dependence on accurate 3D orientation estimates (e.g., roll, pitch, yaw; quaternion; rotation matrix) to properly convert sensor-frame measurements to a global reference frame. Small errors in this component can result in substantial localization errors that have limited the feasibility of inertial pedestrian localization \citep{muse}. Since orientation estimation plays a central role in inertial odometry, we hypothesize that improvements in 3D orientation estimation will result in improvements to localization performance.

\begin{figure}[t]
    \centering
    \includegraphics[width=0.7\linewidth]{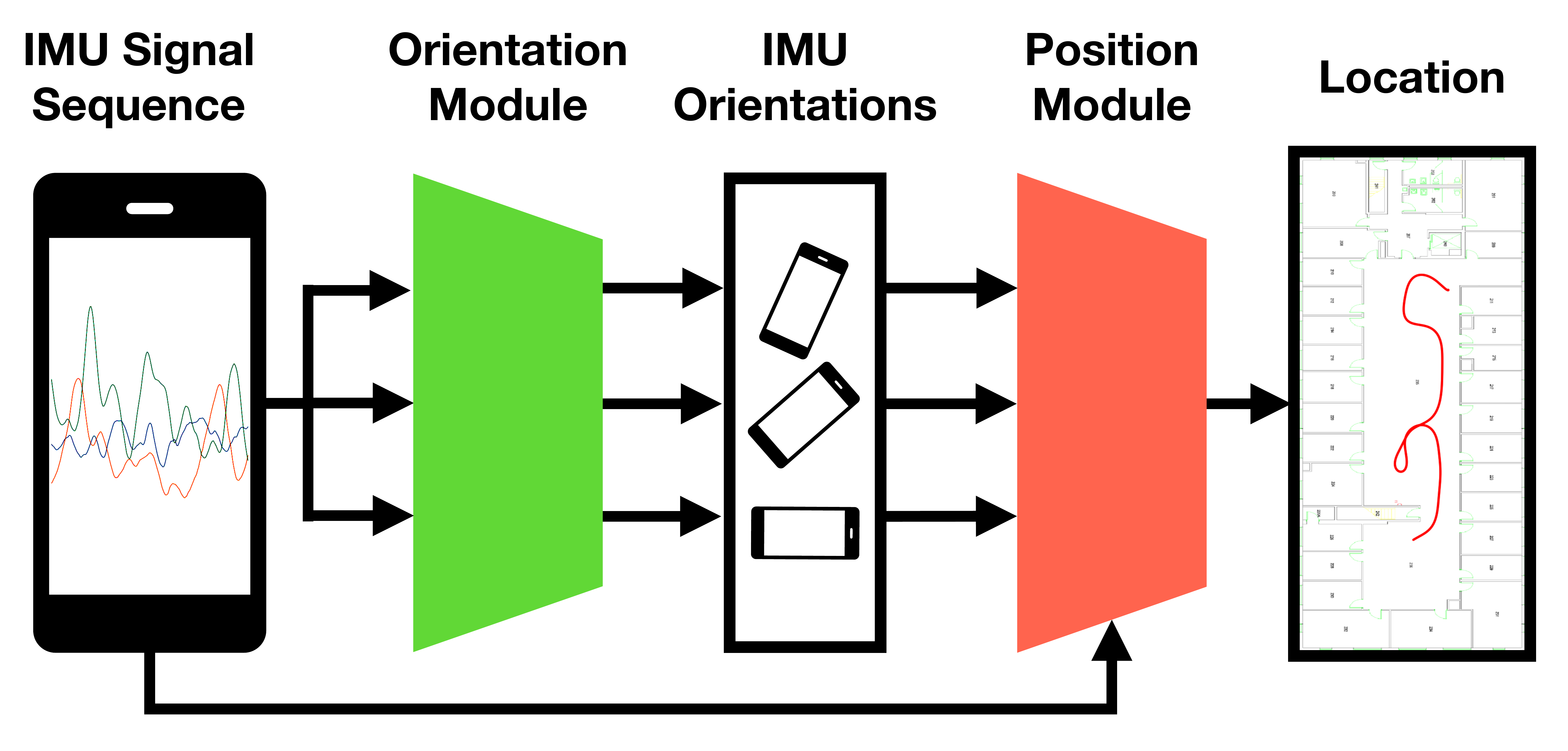}
    \caption[System overview]{Overview of our proposed method.} 
    \label{fig:overview}
\end{figure}

Existing localization methods typically rely on WiFi, Bluetooth, LiDAR, or camera sensors because of their effectiveness. However, WiFi and Bluetooth beacon-based solutions are costly due to requiring heavy instrumentation of the environment for accurate localization \citep{navcog}. While LiDAR-based localization is highly accurate, it is expensive and power-hungry \cite{loam}. Image-based localization is effective with ample light and texture, but is also power-intensive and a privacy concern. IMUs would address many of these problems because they are infrastructure-free, highly energy-efficient, do not require line-of-sight with the environment, and are highly ubiquitous by function of their cost and size.

Recent deep-learning approaches, like IONet \cite{ionet} and RoNIN \cite{ronin}, have demonstrated the possibility of estimating 3D device (or user) motion using an IMU, but they do not directly address device orientation estimation. These new approaches have been able to address the problem of drift suffered by traditional inertial localization techniques through the use of supervised learning to directly estimate the spatial displacement of the device. However, most existing works use the 3D orientation estimates generated by the device (typically with conventional filtering-based approaches), which can be inaccurate ($>$20$^\circ$ of error with the iPhone 8). This orientation is typically used as an initial step to rotate the local IMU measurements to a common reference frame before applying a deep network \cite{ionet}. This is flawed as the deep network output can be corrupted by these orientation estimate errors, leading to significant error growth.

Our approach to this problem involves designing a supervised deep network architecture with an explicit orientation estimation module to complement a position estimation module, shown in Figure \ref{fig:overview}. In addition to the gyroscope, the 3D orientation module makes use of information encoded in the accelerometer and magnetometer of the IMU by proxy of their measurements of gravitational acceleration and the Earth's magnetic field. By looking at a small temporal window of IMU measurements, this module learns to estimate a more accurate device orientation, and by extension, results in a more accurate device location. We train a two-stage deep network to estimate the 5D device pose -- 3D orientation and 2D position (3D position is also possible).

The contributions of our work are: (i) a state-of-the-art deep network architecture to perform inertial orientation estimation, which we show leads to improved position estimates; (ii) an end-to-end model that produces more accurate position estimation than previous classical and learning-based techniques; and (iii) a building-scale inertial sensor dataset of annotated 6D pose (3D position + orientation) during human motion.

\section{Related Work}
We place inertial systems into two broad categories based on their approaches to localization and orientation: traditional methods and data-driven methods.

\subsection{Traditional Localization Methods:} Dead reckoning with an IMU using the analytical solution consists of integrating gyroscopic readings to determine sensor orientation (e.g.,  Rodrigues' rotation formula), using those orientations to rotate accelerometer readings into the global frame, removing gravitational acceleration, and then double-integrating the corrected accelerometer readings to determine position \cite{ionet}. The multiple integrations lead to errors being magnified over time, resulting in an unusable estimate within seconds. However, additional system constraints on sensor placement and movement can be used to reduce the amount of drift, e.g., foot mounted ZUPT inertial navigation systems that rely on the foot being stationary when in contact with the ground to reset errors \cite{PDR_summary}. Extended Kalman Filters (EKFs) are often used to combine IMU readings accurate in the near-term with other localization methods that are more accurate over the long term, like GPS \citep{gps_imu_ekf}, cameras \citep{vins-mono}, and heuristic constraints \citep{handheld_io}.

\subsection{Data-driven Localization Methods:} 
Recent years have seen the development of new data-driven methods for inertial odometry. Earlier work vary from approaches like RIDI \citep{ridi}, which relies on SVMs, to deep network approaches like \citet{speed_estimation}, IONet \citep{ionet}, RoNIN \citep{ronin}, and TLIO \citep{tlio}. 

RIDI and \citet{speed_estimation} fundamentally rely on dead-reckoning, i.e., double-integrating acceleration, but differ in how they counteract the resulting extreme drift. RIDI uses a two-stage system to regress low-frequency corrections to the acceleration. A Support Vector Machine (SVM) classifies the motion as one of four holding modalities (e.g. in the hand, the purse) from the accelerometer/gyroscope measurements. These measurements are then fed to a modality-specific Support Vector Regressor (SVR) to determine a correction applied before the acceleration is double-integrated to determine user position. \citet{speed_estimation} use a convolutional neural network (CNN) to regress momentary user speed as a constraint. The speed acts as a pseudo-measurement update to an EKF that performs accelerometer double-integration as its process update.

IONet and the remaining work forgo dead-reckoning and instead rely on deep networks to bypass one set of integration, thereby limiting error growth. In IONet, a bi-directional long-short-term memory (BiLSTM) network is given a window of world-frame accelerometer and gyroscope measurements (with the reference frame conversion done by the phone API), from which it sequentially regresses a polar displacement vector describing the device's motion in the ground plane. This single integration helps minimize the error magnification. The bulk of their work is evaluated in a small Vicon motion capture studio with different motion modalities, e.g., in the pocket. 

RoNIN builds on the IONet approach and presents three different neural network architectures to tackle inertial localization: an LSTM network, a temporal convolutional network (TCN), and a residual network (ResNet). These models regress user velocity/displacement estimates in the ground (x-y) plane.

TLIO is a recent work that uses a ResNet-style architecture from RoNIN to estimate positional displacements. They fuse these deep estimates with the raw IMU measurements using a stochastic-cloning EKF, which estimates position, orientation, and the sensor biases.

The present work suggests current data-driven inertial localization approaches lack a robust device orientation estimator. Previous networks rely heavily on direct gyroscope integration or the device's estimate, which fuses accelerometer, gyro, and magnetometer readings using classical methods. While these estimates may be accurate over the short term, they are prone to discontinuities, unstable readings, and drift over time. The success of data-driven approaches in localization suggests similar possibilities for orientation estimation.

\subsection{Traditional Orientation Estimation Methods:}
Prior work for device orientation estimation are primarily based on traditional filtering techniques. The Madgwick filter \cite{madgwick} is used widely in robotics. In the Madgwick filter, gyroscope readings are integrated over time to determine an orientation estimate. This is accurate in the short term but drifts due to gyroscope bias. To correct the bias, a minimization is performed between two vectors: (1) the static world gravity vector, rotated into the device frame using the current estimated orientation, and (2) the acceleration vector. The major component of the acceleration vector is assumed to be gravity, so it calculates a gradient to bring the gravity vector closer to the acceleration vector in the current frame. The orientation estimate consists of a weighted combination of this gradient and the gyroscope integration. This assumes the non-gravitational acceleration components are small, which is impractical for pedestrian motion.

Complementary filters are also used in state-of-the-art orientation estimation systems like MUSE \cite{muse}. MUSE behaves similarly to the Madgwick filter, but uses the acceleration vector as the target of the orientation update only when the device is static. Instead, they mainly use the magnetic north vector as the basis of the gradient calculation. This has the advantage of removing the issue of large non-gravitational accelerations causing erroneous updates since when the device is static, the acceleration vector consists mostly of gravity. However, a static device is rare during pedestrian motion and magnetic fields can vary significantly from within the same building due to local disturbances which are difficult to characterize.

Extended Kalman Filter (EKF) approaches \cite{ekf, ekf_iteration, other_ekf} follow a similar approach to the previously mentioned filters, but use a more statistically rigorous method of combining gyroscope integration with accelerometer/magnetometer observations. An estimate of the orientation error can also be extracted from this type of filter. We take advantage of such a filter in our work, but replace the gravity vector or magnetic north measurement update with the output of a learned model to provide a less noisy estimate of the true orientation and simplify the Kalman update equations.

\subsection{Data-driven Orientation Estimation Methods:}
Recent literature like OriNet \citep{orinet} and \citet{denoise_imu} (abbreviated \citetalias{denoise_imu}) has begun utilizing deep networks to regress orientation from IMU measurements. OriNet uses a recurrent neural architecture based on LSTMs to propagate state. It corrects for gyroscopic bias via a genetic algorithm and sensor noise via additive Gaussian noise during training. \citetalias{denoise_imu} estimates orientation via gyroscopic integration, but uses a CNN to perform a correction to the angular velocity to filter out unwanted noise and bias prior to integration. These methods have primarily focused on filtering gyroscopic data using deep networks and estimating correction factors to reduce bias and noise. Our method directly estimates an orientation from all IMU channels using a deep network to capture all error sources for long-term accuracy while fusing gyro data in the short term via an EKF. The prior data-driven approaches thus far have yet to include magnetic observations, which leaves performance on the table given the success of incorporating magnetic observations in classical approaches.

\begin{figure*}
    \centering
    \includegraphics[width=0.95\linewidth]{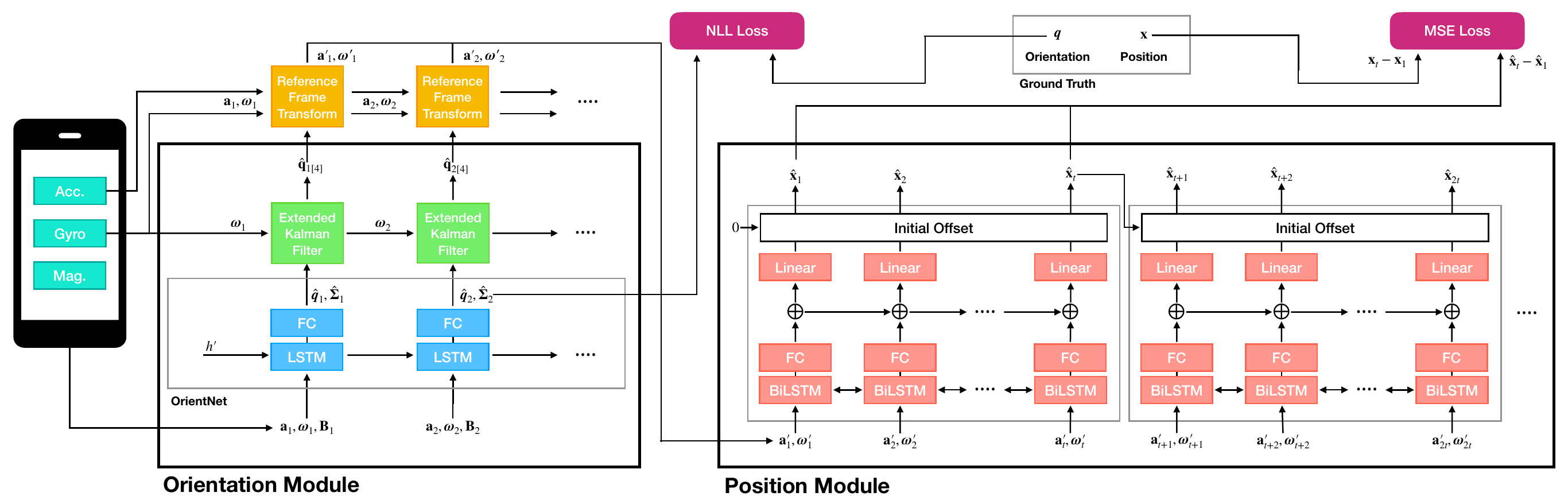}
    \caption[Detailed system diagram]{Detailed system diagram. IMU readings are first passed to the orientation module, which is trained to estimate the orientation quaternion $q$. This orientation is used to convert accelerometer/gyroscope readings from device to world frame. These readings are passed to the position module, which is trained to minimize displacement error per window, for localization.}
    \label{fig:detailed_sys_diagram}
\end{figure*}

\section{Method}
We aim to develop a method for 3D orientation and 2D position estimation of a smartphone IMU held by a pedestrian through the use of supervised learning. Our model is designed based on the knowledge that the accelerometer contains information about the gravitational acceleration of the Earth and that the magnetometer contains information about the Earth's magnetic field. Thus, it should be possible to infer the absolute 3D orientation of the device using a deep network with higher accuracy than that achievable with the heuristic-based traditional filtering methods.

\subsection{Estimating 3D Orientation:}
We propose a network architecture for estimating device orientation. The network consists of two components: (1) an orientation network that estimates a device orientation from the provided acceleration, angular velocity, and magnetometer readings and (2) an Extended Kalman filter to further stabilize the network output with the gyroscope readings. The resulting 3D orientation is used to rotate the accelerometer and gyroscope channels from the phone's coordinate system to a world coordinate system. The corrected measurements are then passed as inputs to the position network. 

We use a neural network, referred to as the Orientation Network (OrientNet), to convert IMU measurements to a 3D orientation and a corresponding covariance estimate. Instead of directly converting the magnetic field or acceleration vector to orientation, as is done in traditional filtering methods, we use a neural network to learn a data-driven mapping of other sensor measurements to orientation. We find that the magnetic field measurements contribute most reliably to this estimate (much more than gravity), in agreement with the claims by \citet{muse}.

Formally, we estimated the instantaneous 3D orientation 
\begin{align}
\hat{\boldsymbol{\theta}}, \hat{\boldsymbol{\Sigma}} = g(\mathbf{a}_t,\boldsymbol{\omega}_t, \mathbf{B}_t, \mathbf{h}'_{t-1}), 
\end{align}
where the function $g$ consists of a 2-layer LSTM with 100 hidden units and $\mathbf{h}'_{t-1}$ is a hidden state produced by the LSTM at the last time step. At each time step, the accelerometer, gyroscope, and magnetometer readings ($\mathbf{a}, \boldsymbol{\omega}, \mathbf{B}$) are taken as input. The hidden state is then fed through 2 fully-connected layers to produce an absolute orientation $\hat{\boldsymbol{\theta}}$ in the global reference frame, and through two other fully connected layers to produce an orientation covariance $\boldsymbol{\hat{\Sigma}}$. This covariance represents the auto-covariance of the 3-dim orientation error, determined using a boxminus operation between the true and estimated orientations (as defined in equation \ref{eqn:manifold}), and is thus a 3$\times$3 matrix.  In the position estimation network described later, this orientation estimate will be used as a coordinate transform to rotate the IMU channels from the local phone frame to the global reference frame. 

We find that the OrientNet maintains high accuracy in its orientation estimate over long periods of time, but does not achieve the fine-grain accuracy of gyroscope integration. Using the raw gyroscope measurements as a process update and the outputs of the orientation estimation network as a measurement update in an Extended Kalman Filter (EKF), we can achieve higher local and global accuracy. We find the EKF outperforms deep networks at performing this fusion as it handles the angular data with well-defined quaternion operations while allowing for stable and intuitive fusion of our OrientNet outputs. We use a quaternion EKF \cite{ekf} with process updates defined by
\begin{equation}
    \begin{aligned}
        \mathbf{\hat{x}}_{k|k-1} &= \mathbf{\hat{x}}_{k-1|k-1} + \mathbf{B}_k\boldsymbol{\omega}_{k}, \\
        \mathbf{\hat{P}}_{k|k-1} &=  \mathbf{\hat{P}}_{k-1|k-1}  + \mathbf{Q}_k,
    \end{aligned}
\end{equation}
where the quaternion state $\mathbf{\hat{x}}_{k-1|k-1}$ is the \textit{a posteriori} estimate of of the state at $k-1$ given observations up to $k-1$, and $\mathbf{\hat{x}}_{k|k-1}$ is the propagated orientation estimate at timestep $k$ given observations up to $k-1$. The motion model is parameterized by $\mathbf{B}_k$, which converts the current gyroscope measurement $\boldsymbol{\omega}_k$ into a quaternion representing the rotation achieved by $\boldsymbol{\omega}_k$. The process update is applied via simple addition, which approximates quaternion rotations at high sample rates. $\mathbf{\hat{P}}_{k|k-1}$ is the estimate of the covariance of the propagated state vector at time $k$ given observations up to $k-1$, with $\mathbf{\hat{P}}_{k-1|k-1}$ again the \textit{a posteriori} estimate of covariance at time $k-1$. $\mathbf{Q_k}$ is a static diagonal propagation noise matrix for the gyro, which we set to $0.005\mathcal{I}_3$ based on experimentation with our training data.

The EKF's measurement updates correct the propagated state with the network-predicted orientation. Using normal addition and subtraction as orientation operators becomes inaccurate since there is no guarantee the predicted and propagated quaternions lie close together. Thus, here we treat the difference between orientations as a distance on the quaternion manifold instead of as a vector-space distance using the methods presented by \citet{hertzberg_integrating_2011}. Boxplus ($\boxplus$) and boxminus ($\boxminus$), which respect the manifold, replace addition and subtraction of quaternions:
\begin{equation}
    \begin{aligned}
        \mathbf{q}_1 \boxminus \mathbf{q}_2 &= 2\bar{\log}(\mathbf{q}_2^{-1}\otimes \mathbf{q}_1) = \boldsymbol{\delta}, \\
        \mathbf{q}_1 \boxplus \boldsymbol{\delta} &= \mathbf{q}_1 \otimes \exp(\boldsymbol{\delta}/2) = \mathbf{q}_2, \\
        \exp(\boldsymbol{\delta}) &= \begin{bmatrix} \cos(\|\boldsymbol{\delta}\|)\\ \textrm{sinc}(\|\boldsymbol{\delta}\|)\boldsymbol{\delta} \end{bmatrix}, \\
        \bar{\log}(\begin{bmatrix} w \\ \mathbf{v} \end{bmatrix}) &= \begin{cases}
          0, &  \mathbf{v} = 0 \\
          \frac{\textrm{atan}(\|\mathbf{v}\|/w)}{\|\mathbf{v}\|}\mathbf{v}, & \mathbf{v} \neq 0
        \end{cases}.
    \end{aligned}
    \label{eqn:manifold}
\end{equation}
$\mathbf{q}_1$ and $\mathbf{q}_2$ are unit quaternions; $\boldsymbol{\delta}$ is the three-dimensional manifold difference between them. $w$ and $\mathbf{v}$ are the real and imaginary parts of a quaternion, respectively. These operators maintain the unit norm and validity of the resulting quaternions. The norm of $\boldsymbol{\delta}$ between two quaternions describes the distance along a unit sphere between the orientations. Adding $\boldsymbol{\delta}$ to a quaternion using $\boxplus$ results in another valid quaternion displaced from the initial quaternion. This displacement is encoded in $\boldsymbol{\delta}$. With these operators, the measurement update for our method becomes
\begin{equation}
    \begin{aligned}
        \mathbf{K}_k &=  \mathbf{P}_{k|k-1} (\mathbf{P}_{k|k-1} + \mathbf{R}_k)^{-1},\\
        \mathbf{\hat{x}}_{k|k} &= \mathbf{\hat{x}}_{k|k-1} \boxplus \mathbf{K}_k(\mathbf{q}_k \boxminus  \mathbf{\hat{x}}_{k|k-1}), \\
        \mathbf{P}_{k|k} &= (\mathbf{I} - \mathbf{K}_k)\mathbf{P}_{k|k-1},
    \end{aligned}
\end{equation}
where $\mathbf{q}_k$ and $\mathbf{R}_k$ are the network-predicted orientation and covariance for timestep $k$. The result of the measurement update is the final orientation estimate for timestep $k$ ($\mathbf{\hat{x}}_{k|k}$) and the estimated state covariance $\mathbf{P}_{k|k}$.

\subsection{Orientation Module Training:}
To train the orientation module, we first perform a traditional ellipsoid fit calibration on the raw magnetometer values, which also serves to scale the network inputs to a confined range \cite{mag_cali}. From here on, we will refer to these coarsely calibrated magnetometer readings as part of the raw IMU measurements. To obtain the mean and covariance needed to parameterize a Gaussian estimate, a negative log likelihood (NLL) loss is used to train the covariance estimator for each orientation output. This loss seeks to maximize the probability of the given ground truth, assuming a Gaussian distribution parameterized by the estimated orientation and covariance:
\begin{equation}
    \mathcal{L}_{\text{orient}} = \frac{1}{2}(\boldsymbol{q} \boxminus \hat{\boldsymbol{q}})^T \hat{\boldsymbol\Sigma}^{-1} (\boldsymbol{q} \boxminus \hat{\boldsymbol{q}}) + \frac{1}{2}\ln(|\hat{\boldsymbol\Sigma}|).
    \label{eqn:orient_loss_mle}
\end{equation}
The output of the covariance estimation head of the network is a six dimensional vector, following a standard parameterization of a covariance matrix \cite{russell_multivariate}. The first three elements are the log of the standard deviation, which are then exponentiated and squared to form the diagonal elements of the covariance matrix. The remaining 3 are correlation coefficients between variables, which are multiplied by the relevant exponentiated standard deviations to form the covariance elements of the matrix. 

\subsection{Estimating 2D Position:}
Analytically, to convert from the world frame accelerometer values to position, one would need to perform two integrals. However, any offsets in the acceleration values would result in quadratic error growth. Therefore, we again adopt the use of a neural network to learn an appropriate approximation that is less susceptible to error accumulation, as has been demonstrated successfully by \citet{ronin} and \citet{ionet}. The position estimation network takes world frame gyroscope and accelerometer channels as inputs and outputs the final user position in the same global reference frame as the orientation module. We use a Cartesian parameterization of the user position to match that of the rotated accelerometer.

The position module's architecture is depicted in Figure \ref{fig:detailed_sys_diagram}. We primarily rely on a 2-layer BiLSTM with a hidden size of 100. The input is a sequence of 6-DOF IMU measurements in the world frame. At each timestep, the hidden state is passed to 2 fully-connected layers with $\tanh$ activation between them and hidden state sizes of 50 and 20, respectively. The resulting vector is then passed to a linear layer that converts it to a two-dimensional Cartesian displacement relative to the start of each window. These are then summed over time to form Cartesian positions relative to the start of the sequence, with each window's final position serving as the initial offset for the next window. During test time, the LSTM hidden states are \textit{not} propagated between each sequence window, as this periodic resetting helps to limit the accumulation of drift.

\subsection{Position Module Training:}
Over the course of training, progressively longer batch sequences are provided. We start with sequences of length 100 and progressively increase this over training to length 2000. We find this type of curriculum learning greatly reduces drift accumulation as the overall error must be kept low over a longer time period. After this routine, the sequence length may then be dropped back down to a shorter length to reduce latency. We use MSE loss over the displacement of each LSTM window. In other words,
\begin{equation}
    \mathcal{L}_\text{position} = \mathcal{L}_{\text{MSE}}(\mathbf{x}_t - \mathbf{x}_1, \hat{\mathbf{x}}_t - \hat{\mathbf{x}}_1)\,.
\end{equation}

\section{Dataset}
To collect trajectories through the narrow hallways of a typical building, we rely on a SLAM rig (Kaarta Stencil) as ground truth. To obtain the phone's ground truth orientation and position, we rigidly mount it to the rig, which uses a LiDAR sensor, video camera, and Xsens IMU to estimate its pose at 200Hz. From testing in a Vicon motion capture studio, we measured $<$1.5$^\circ$ RMS orientation error and $<$10cm RMS position error. Given the position error is less than the size of most smartphones, the Stencil is accurate enough to serve as ground truth while having the advantage of not being constrained to a single room. Based on this testing, we also apply low pass filtering to the trajectory so that it agrees better with the Vicon output. Smartphone data is collected using an iPhone 8, from which we obtain raw accelerometer, gyroscope, and magnetometer readings at 100Hz. Most trajectories in this dataset are about 10 minutes in length. We collected 20 hours of human motion in 3 different buildings of varying shapes and sizes. 15 users of different physical builds were instructed to walk with variable speeds, pauses, and arbitrary directional changes while carrying the mapping rig and smartphone. Participants were allowed to hold the rig however they wished (e.g., 90 degrees offset) and to readjust their grip as needed, so trajectories include variations in rig orientation relative to the user. An initial magnetometer calibration \cite{mag_cali} was performed at the start location.

\begin{figure}[t]
    \centering
    \includegraphics[width=0.5\linewidth]{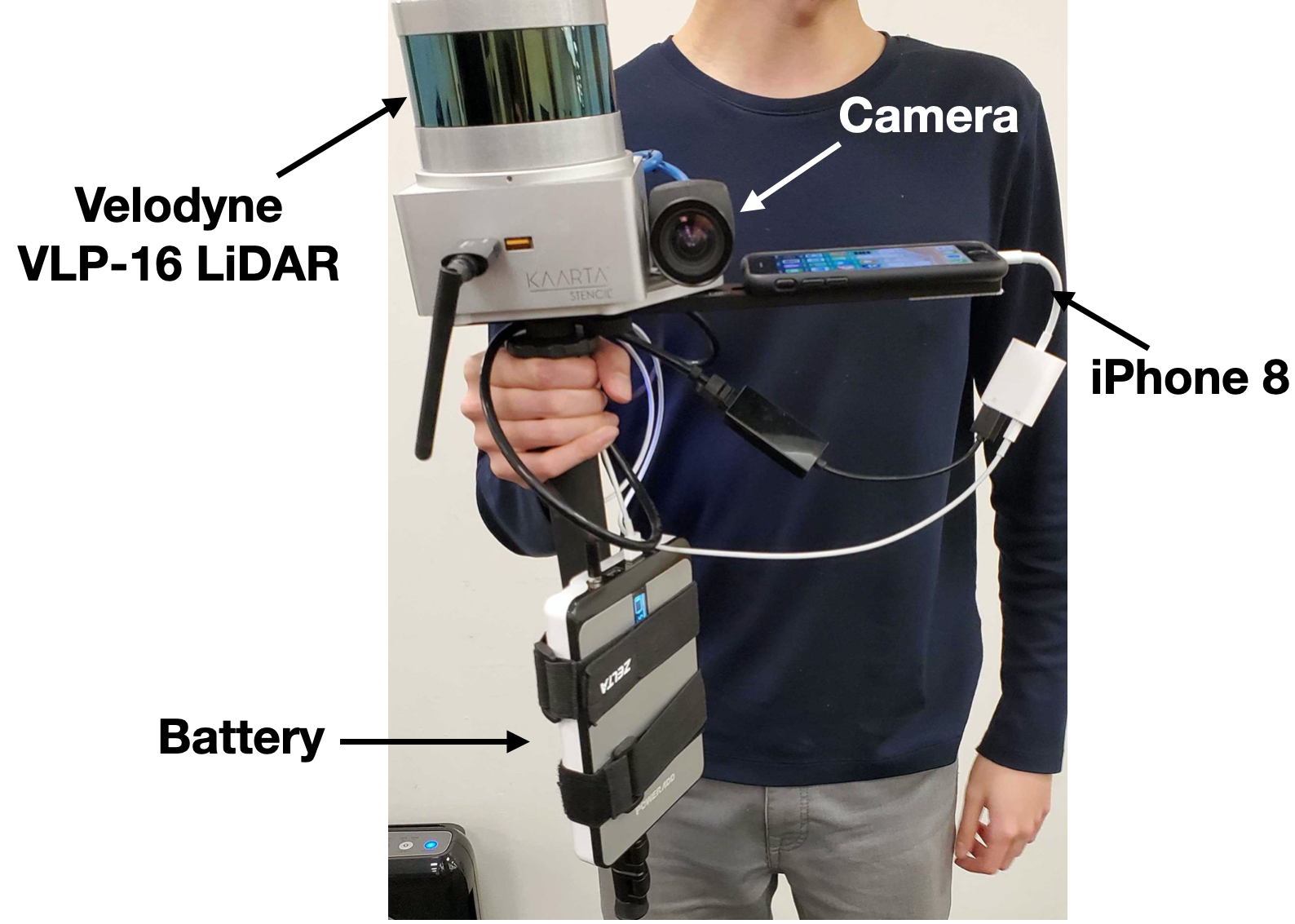}
    \caption[Data collection rig]{Data collection SLAM rig (Kaarta Stencil)}
    \label{fig:data_collection}
\end{figure}

Since the user must carry the mapping system to which the phone is mounted, the user's motions are somewhat unnatural. However, we argue that the major factors that complicate inertial pedestrian localization are maintained, such as the lack of both zero-velocity points and movement constraints on any axis of the device. Subjects were allowed to shake and orient the rig relative to their walking direction however they liked. Furthermore, RoNIN and IONet have already demonstrated that deep networks can generalize across different ways of holding a phone and to different brands of smartphones. Because testing such modalities makes it much more difficult to acquire ground truth device orientation (such as in-the-pocket), we primarily rely on the induced motion over the course of normal movement to generate realistic device motions. While RoNIN and IONet provide two of the largest datasets for pedestrian inertial odometry, they lack the necessary data channels for our model. IONet's dataset, OxIOD \cite{oxiod}, lacks raw IMU values without the iOS CoreMotion processing and coordinate transform already applied, so our orientation module is unable to be used. Furthermore, their ground truth orientations display consistent artifacts at certain orientations, which would corrupt a supervised model trained on them. RoNIN does not provide ground truth phone orientations, instead opting to provide the orientation of a second Google Tango phone attached to the user's body; this means we cannot train our orientation module on their trajectories. TLIO did not release their dataset. While there are several visual-inertial datasets, like ADVIO \cite{ADVIO}, they do not provide the relative pose between initializations for each trajectory. Since our method requires that all data be aligned to the same frame (due to the reliance on magnetometer data), we are unable to train or evaluate on these datasets. EuRoC \citep{euroc} and TUMVI \citep{tumvi} suffer from similar problems and do not include magnetometer readings for our models. These datasets also tend to have much shorter trajectories than ours, so drift is less evident. We hope to demonstrate to utility of including such information in future datasets.

\section{Experiments}
To demonstrate the effectiveness and utility of our inertial odometry system, we set three main goals for evaluation: (i) verify that our model produces better orientation estimates than the baselines, (ii) show that our model is able to achieve higher position localization accuracy than previous methods, and (iii) demonstrate that orientation error is a major source of final position error by showing that other inertial odometry methods benefit from our orientation module. 

The main metric used to evaluate the orientation module is root mean squared (RMSE) orientation error, measured as the direct angular distance between the estimated and ground truth orientations. We evaluate the accuracy of our position estimate using metrics defined by \citet{sturm11rss-rgbd} and used by RoNIN:
\begin{itemize}
    \item \textbf{Absolute Trajectory Error (ATE)}: the RMSE between corresponding points in the estimated and ground truth trajectories. The error is defined as $E_i = \boldsymbol{x}_i - \hat{\boldsymbol{x}}_i$ where $i$ corresponds to the timestep. This is a measure of global consistency and usually increases with trajectory length.
    \item \textbf{Time-Normalized Relative Traj. Error (T-RTE)}: the RMSE between the displacements over all corresponding 1-minute windows in the estimated and ground truth trajectories. The error is defined as $E_i = (\boldsymbol{x}_{i+t} - \boldsymbol{x}_i) - (\hat{\boldsymbol{x}}_{i+t} - \hat{\boldsymbol{x}}_i)$ where $i$ is the timestep and $t$ is the interval. This measures local consistency between trajectories.
    \item \textbf{Distance-Normalized Relative Traj. Error (D-RTE)}: the RMSE between the displacements over all corresponding windows in the estimated and ground truth trajectories where the ground truth trajectory has traveled 1 meter. The error is defined as $E_i = (\boldsymbol{x}_{i+t_d} - \boldsymbol{x}_i) - (\hat{\boldsymbol{x}}_{i+t_d} - \hat{\boldsymbol{x}}_i)$ where $i$ corresponds to the timestep and $t_d$ is the interval length required to traverse a distance of 1m.
\end{itemize}
The RMSE for these metrics is calculated using the following equation where $E_i$ is the $i$-th error term out of $m$ total:
\begin{equation}
    RMSE = \sqrt{\frac{1}{m} \sum_{i=1}^m \Vert E_i \Vert_2^2}.
    \label{eqn:rmse}
\end{equation}

\subsection{Training/Testing:}
We implemented our model in Pytorch 1.15 \cite{pytorch} and train it using the Adam optimizer \cite{adam} on an Nvidia RTX 2080Ti GPU. The orientation network is first individually trained using a fixed seed and a learning rate of 0.0005. Then, using these initialized weights, the position network is attached and then trained using a learning rate of 0.001. We use a batch size of 64, with the network reaching convergence within 20 epochs. Each epoch involves a full pass through all training data. 

At test time, an initial orientation can be provided or, assuming a calibrated magnetometer, the initial orientation can be estimated by the network directly with high accuracy relative to a predefined global frame. This cannot be said for systems that rely solely on gyroscope integration, which produces an orientation relative to the initialization. As this system is meant to aid pedestrian navigation using a smartphone IMU, it must have limited computational demands. Using only an AMD Ryzen Threadripper 1920x CPU, the forward inference time is approx. 65ms for 1s of data (100 samples), which suggests real-time capabilities on mobile processors. 

\subsection{Baselines:}
To evaluate our orientation module, we compare it against the iOS CoreMotion API, \citetalias{denoise_imu}, and MUSE \cite{muse}. The CoreMotion estimate is selected for its ubiquity; \citetalias{denoise_imu} is the most competitive deep learning estimator since they outperform OriNet; MUSE is a high-performance traditional approach. As a reminder, CoreMotion and MUSE fuse magnetic readings.

To show the performance of our inertial odometry pipeline, we compare it against several different baseline inertial odometry methods. Pedestrian Dead Reckoning is chosen as the representative of traditional odometry methods. We use a similar PDR baseline to \cite{ronin} that involves regressing a heading and distance every physical step. We assume a stride length of 0.67m/step and use the heading from iOS CoreMotion.

The main data-driven inertial localization methods explored in prior work are IONet, RoNIN, and TLIO, all of which take orientation estimates directly from the phone API. For IONet, we use our own implementation as the original code is not publicly available. IONet was primarily evaluated in a small Vicon motion capture room. We have found, however, that IONet does not perform very well in large indoor environments, which is consistent with experiments run by \citet{ronin}. We evaluate all 3 RoNIN variants--LSTM, TCN, and ResNet--using their exact open source implementation. In our evaluations using their code, we noticed a bug in their evaluation metric, where they omitted the L2-norm in their calculation of RMSE when deriving ATE and RTE (see Equation \ref{eqn:rmse}). Because of this error, their metrics consistently under-report the true error; however, the relative comparisons between their models and the conclusions are still valid because this is applied consistently. We use the correct method for these metrics, which explains the discrepancies between the relative sizes of our errors (in addition to trajectories being from different buildings and of different lengths). TLIO uses RoNIN-ResNet with stochastic-cloning EKF to refine the orientation estimates; we use their released code for evaluation.

\begin{table}
    \footnotesize
    \centering
    \begin{tabular}{lrrr}
        \toprule
        \textbf{System} & \textbf{Bldg 1} & \textbf{Bldg 2} & \textbf{Bldg 3} \\
        \midrule
        iOS CoreMotion & 0.39 & 0.37 & 0.40 \\
        MUSE & 0.21 & 0.25 & 0.45 \\
        \citetalias{denoise_imu} & 0.23 & 0.30 & 0.47 \\
        OrientNet only (ours) & 0.21 & 0.44 & 0.49 \\
        OrientNet+EKF (ours) & \textbf{0.08} & \textbf{0.10} & \textbf{0.14} \\
        \bottomrule
    \end{tabular}
    \caption[Orientation angular RMSE comparison]{Orientation RMSE comparison (in radians). Each building is separately trained and tested; building test sets are of similar length ($\sim$2.5 hr each).}
    \label{tab:ios_vs_orientnet}
\end{table}

\begin{figure}
    \centering
    \includegraphics[width=\linewidth]{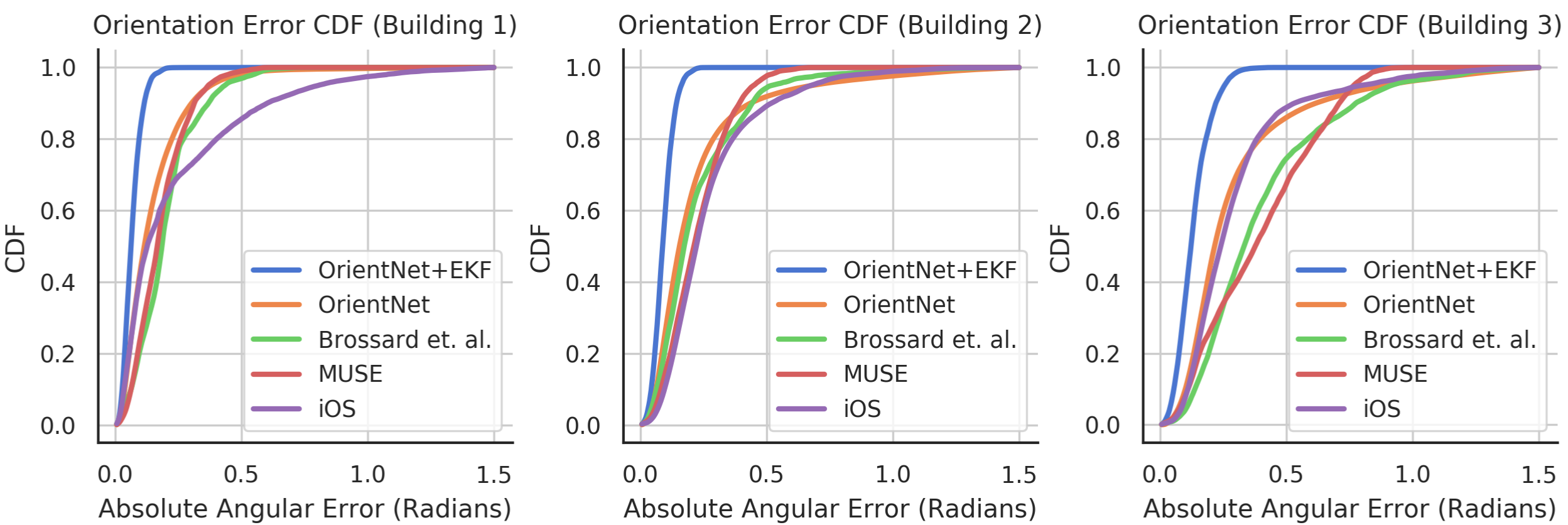}
    \caption{Comparison of model CDFs for orientation error}
    \label{fig:orient_cdf}
\end{figure}

\subsection{Orientation Analysis:}
We now seek to answer the question of whether our orientation pipeline is worth using, i.e., does it outperform the systems that others use? We perform evaluations on our building-scale dataset, where each trajectory occurs over a long time period of 10 minutes that allows one to more easily discern accumulated drift. Table \ref{tab:ios_vs_orientnet} demonstrates that our model outperforms competing approaches by a considerable margin when trained separately on trajectories from each building. Averaging across all three buildings, our estimate is 0.28 radians (16.04$^\circ$) more accurate than CoreMotion's estimate, 0.22 radians (12.83$^\circ$) more accurate than \citetalias{denoise_imu}, and 0.20 radians (11.50$^\circ$) more accurate than MUSE's. While MUSE and \citetalias{denoise_imu} outperform the base OrientNet slightly, the OrientNet+EKF maintains a significant lead. In fact, at 0.08 radians (4.6$^\circ$) in Bldg 1, our method nearly reaches the ground truth rig's accuracy.

Figure \ref{fig:orient_cdf}'s comparison of the error CDFs is particularly useful in understanding relative performance. We can see that the EKF addresses one of the main limitations of using only the OrientNet--namely that the outliers with high errors are eliminated. While the base OrientNet performs better than the other approaches most of the time, it has a larger proportion of large error terms due to jitter and occasional discontinuities that appear in the output, which reduces RMSE down to the level of the others. Our full orientation module significantly outperforms other methods in all metrics with better performance and fewer outliers.

\begin{figure}[t]
    \centering
    \begin{subfigure}[t]{.47\linewidth}
      \centering
      \includegraphics[width=\linewidth]{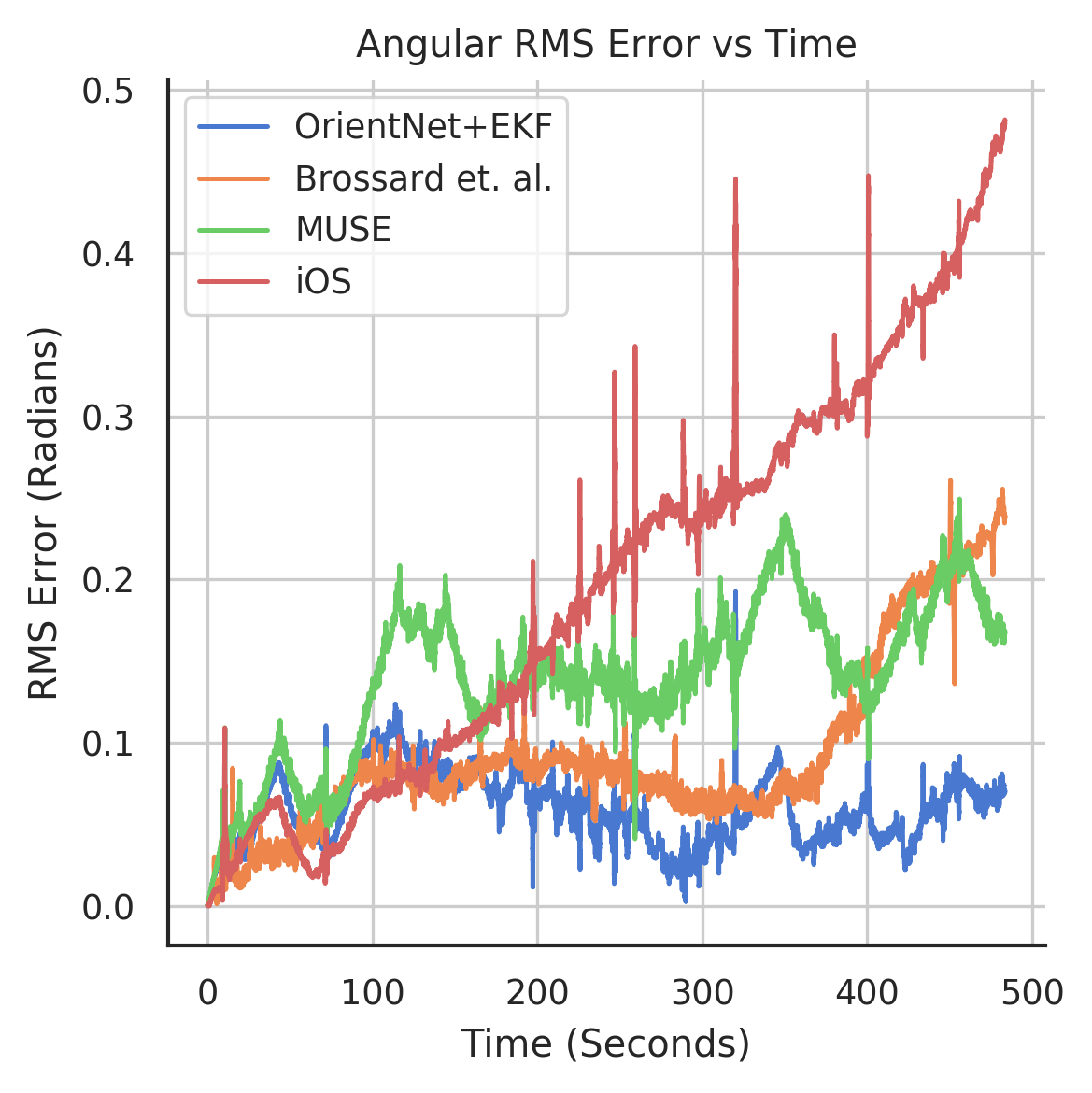} 
      \caption{Model error between true and predicted orientations.}
      \label{fig:orient_error_vis}
    \end{subfigure}
    \begin{subfigure}[t]{.47\linewidth}
      \centering
      \includegraphics[width=\linewidth]{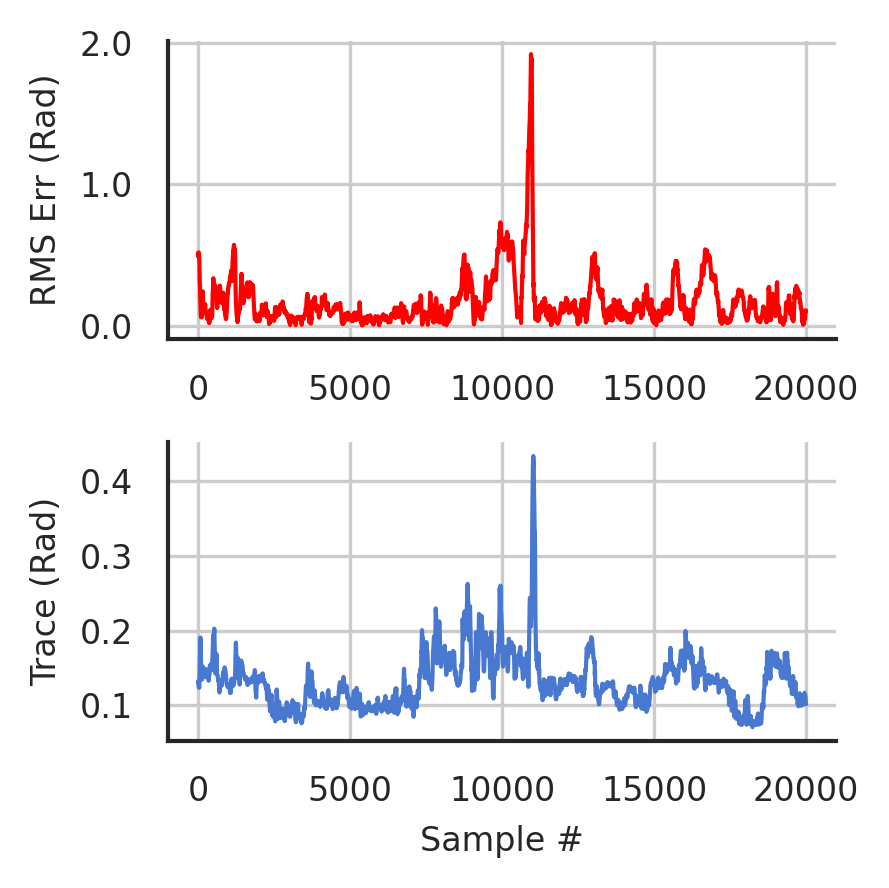}  
      \caption{Correlation between orientation error \& covariance estimate of predicted error dist.}
      \label{fig:orient_cov_vis}
    \end{subfigure}
    \caption{Orientation module performance}
    \label{fig:orient_performance}
\end{figure}

The error growth over time is evident in Figure \ref{fig:orient_error_vis}, where all other methods exhibit a steeper error growth than ours--which stays relatively flat. While MUSE and the iOS estimate can sometimes recover from such drastic error growth eventually via magnetic observations, our pipeline quickly and frequently adapts to keep the orientation estimate accurate in the face of constant device motion. Figure \ref{fig:orient_cov_vis} shows the predicted standard deviation correlates well with the actual error. The square root of the trace of the predicted orientation covariance matrix is, due to the manifold structure of our loss, the standard deviation of the absolute angular error. Overall, 60\% our estimates lie within one predicted standard deviation (a new covariance is predicted for each timestep) of the true orientation, 90\% lie within 2 standard deviations, and 97\% lie within 3. This approximately matches with the expected probabilities of a Gaussian distribution, which suggests our network is producing reasonable covariance estimates.
\begin{table}[t]
    \scriptsize
    \centering
    \begin{tabular}{lrrrrrr}
        \toprule
        \multirow{2}{*}{\textbf{Model}} & \multicolumn{3}{c}{Bldg 1, Known Subjects} & \multicolumn{3}{c}{Bldg 1, Unknown Subjects} \\
        \cmidrule(lr){2-4}
        \cmidrule(lr){5-7}
        & \textbf{ATE} & \textbf{T-RTE} & \textbf{D-RTE} & \textbf{ATE} & \textbf{T-RTE} & \textbf{D-RTE} \\
        \midrule
        PDR & 26.98 &	16.49 &	2.26 & 24.29	& 12.65 &	2.77 \\
        IONet &  33.42 &	22.97 &	2.47 &	31.28 &	24.04 &	2.29  \\
        RoNIN-LSTM     & 18.62 &	7.02 &	0.53 &	18.17 &	6.51 &	0.51 \\
        RoNIN-TCN      & 12.00 &	6.41 &	0.48 &	13.41 &	5.82 &	0.48 \\
        RoNIN-ResNet    & 9.03 &	6.43 &	0.56 &	12.07 &	5.95 &	0.49 \\
        TLIO & 4.62 &	2.52 &	0.31 &	6.34 &	4.22	 & 0.46\\
        Ours & \textbf{4.39} &	\textbf{2.14} &	\textbf{0.30} &	\textbf{5.65} &	\textbf{2.61} &	\textbf{0.38} \\
        \bottomrule
    \end{tabular}
    \caption{Model position generalization across subjects. Known subjects (2.4hr) present in train split; unknown (2.2hr) were not.}
    \label{tab:position_error_subject}
\end{table}
\begin{table}[t]
    \scriptsize
    \centering
    \setlength\tabcolsep{2pt}
    \begin{tabular}{lccccccccc}
        \toprule
        \multirow{2}{*}{\textbf{Model}} & \multicolumn{3}{c}{Building 1} &
        \multicolumn{3}{c}{Building 2} & \multicolumn{3}{c}{Building 3} \\
        \cmidrule(lr){2-4}
        \cmidrule(lr){5-7}
        \cmidrule(lr){8-10}
        & \textbf{ATE} & \textbf{T-RTE} & \textbf{D-RTE} & \textbf{ATE} & \textbf{T-RTE} & \textbf{D-RTE} & \textbf{ATE} & \textbf{T-RTE} & \textbf{D-RTE} \\
        \midrule
        PDR & 25.70 &	14.66 &	2.50 & 21.86	& 19.48 &	1.66 & 12.66	& 12.74 &	1.09 \\
        R-LSTM      &  18.41 &	6.78 &	0.52 &	 29.81 &	18.67 &	0.75 & 33.69 &	13.14 &	0.62 \\
        R-TCN       &  12.67 &	6.13 &	0.48 &	22.52 &	13.69 &	0.73 & 24.79 &	12.48 &	0.59 \\
        R-ResNet    &  10.48 &	6.20 &	0.53 & 35.44	& 15.71 &	0.49 & 14.11 &	11.78 &	0.60 \\
        TLIO    & 5.44 & 3.33 & 0.38 & 8.69	& 8.86 &	\textbf{0.33} & 6.88 &	6.68 &	0.34 \\
        Ours   &  \textbf{4.99} & \textbf{2.37} & \textbf{0.34} & \textbf{8.33} & \textbf{5.97} & 0.41 & \textbf{6.62} & \textbf{2.86} & \textbf{0.26} \\
        \bottomrule
    \end{tabular}
    \caption{Comparison across buildings using separately-trained models. RoNIN models abbreviated with "R-".}
    \label{tab:position_error_building}
\end{table}
\begin{table}[t]
    \scriptsize
    \centering
    \begin{tabular}{llrrrr}
        \toprule
        \textbf{Method} & \textbf{Metric} & \textbf{R-LSTM} & \textbf{R-TCN} & \textbf{R-ResNet} & \textbf{TLIO} \\
        \midrule
        \multirow{3}{*}{\shortstack[l]{API \\ Orientation}} & ATE & 18.41 & 12.67 & 10.48 & 5.44 \\
        & T-RTE & 6.78 & 6.13 & 6.20 & 3.33 \\
        & D-RTE & 0.52 & 0.48 & 0.53 & 0.38 \\
        \midrule
        \multirow{3}{*}{\shortstack[l]{Our \\ Orientation}} & ATE & 7.03 & 6.04 & 5.66	& 4.67 \\
        & T-RTE & 2.71 & 2.56 &	2.63 & 2.39	\\
        & D-RTE & 0.35 & 0.30 & 0.39 & 0.29 \\
         \midrule
        \multirow{3}{*}{\shortstack[l]{True \\ Orientation}} & ATE & 6.53 & 5.69 & 4.49 & 4.53  \\
        & T-RTE & 2.33 & 2.17 & 2.17 & 2.30 \\
        & D-RTE & 0.28 & 0.26 & 0.38 & 0.27 \\
        \bottomrule
    \end{tabular}
    \caption[Comparison of orientation estimators' effects on position networks]{Localization using different orientation estimates on Building 1. RoNIN models abbreviated with "R-".}
    \label{tab:orient_net_vs_gt}
\end{table}

\subsection{Position Analysis:}
Tables \ref{tab:position_error_subject} and \ref{tab:position_error_building} show the comparison between between our end-to-end model and a mix of traditional and deep-learning baselines. Table \ref{tab:position_error_subject} demonstrates that a single model trained on Building 1 generalizes well to both Known Subject and Unknown Subjects test sets; furthermore, it outperforms all other methods on both sets. TLIO is closest in performance because their EKF helps reduce drift by estimating sensor biases. One point to note is the consistent performance of PDR in the ATE metric. It is capable of achieving (relatively) low errors for this metric because of the fewer updates which take place as a result of step counting, so the overall trajectory tends to stay in the same general region. Some of the other models tend to drift slowly over time until the trajectory is no longer centered in the same original location despite almost always producing more accurate trajectory shapes, as reflected by the lower RTE metrics. IONet does not perform well on these large buildings, so will be omitted for the remaining results. 

Table \ref{tab:position_error_building} presents the results of separately training models for evaluation per building. Here, our position estimate outperforms all other methods, especially in RTE. Lower RTE means the trajectory shape is more similar to ground truth while lower ATE means the position has generally deviated less. Note that Bldg 2 and 3 result in larger errors due to their size and the increased presence of magnetic distortions that degrade orientation estimates reliant on magnetic readings. 

\begin{figure}[t]
    \centering
    \includegraphics[width=0.75\linewidth]{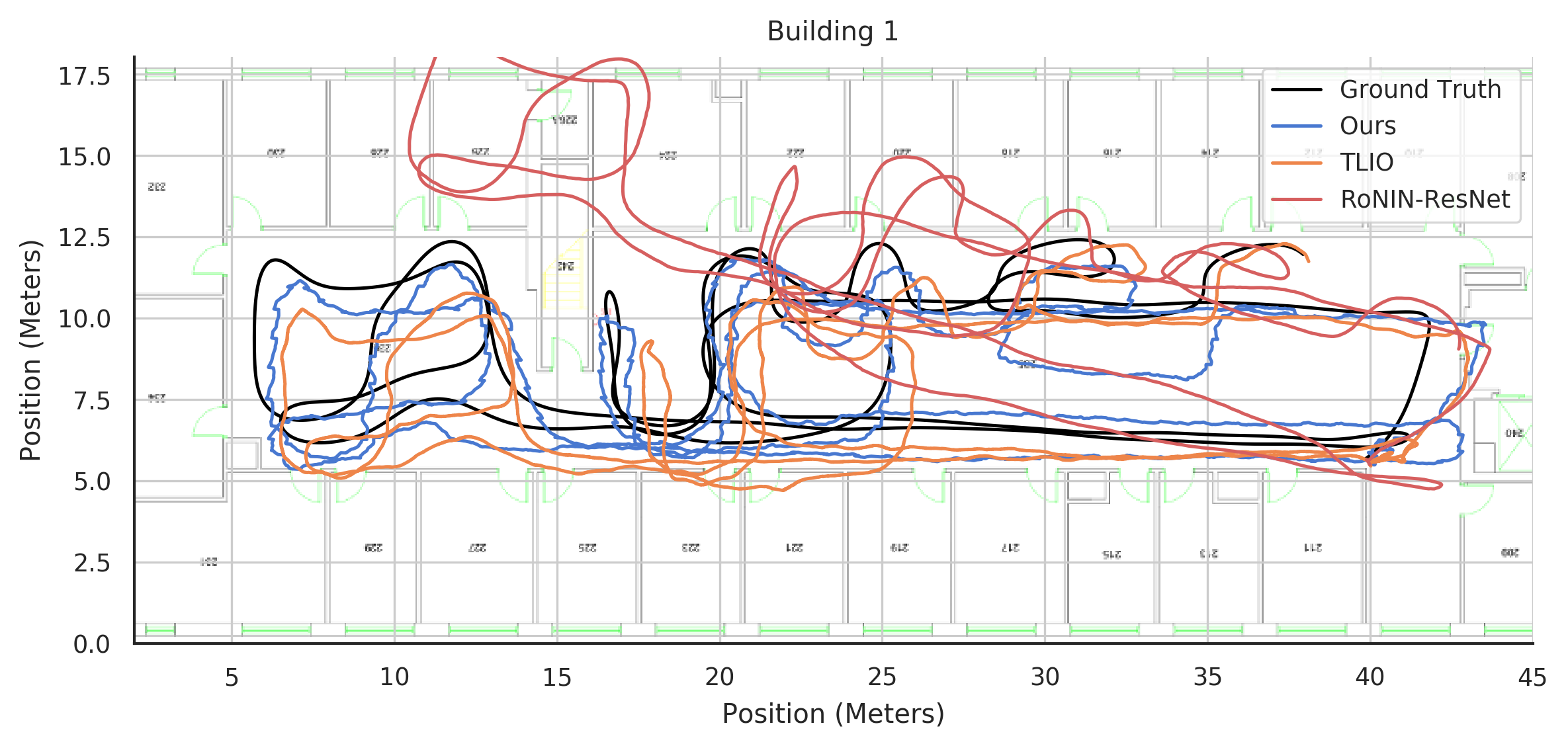}
    \includegraphics[width=0.75\linewidth]{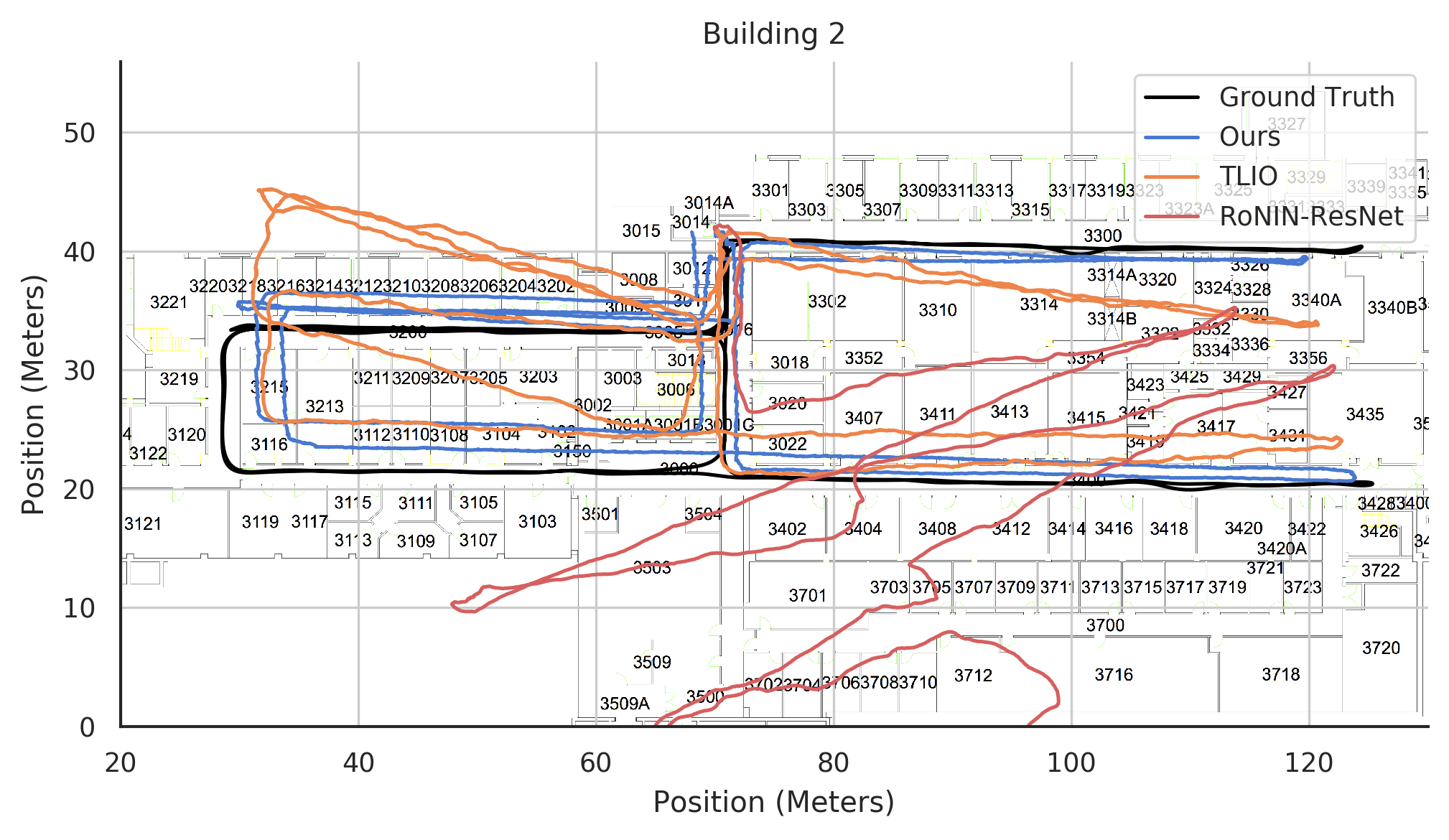}
    \includegraphics[width=0.75\linewidth]{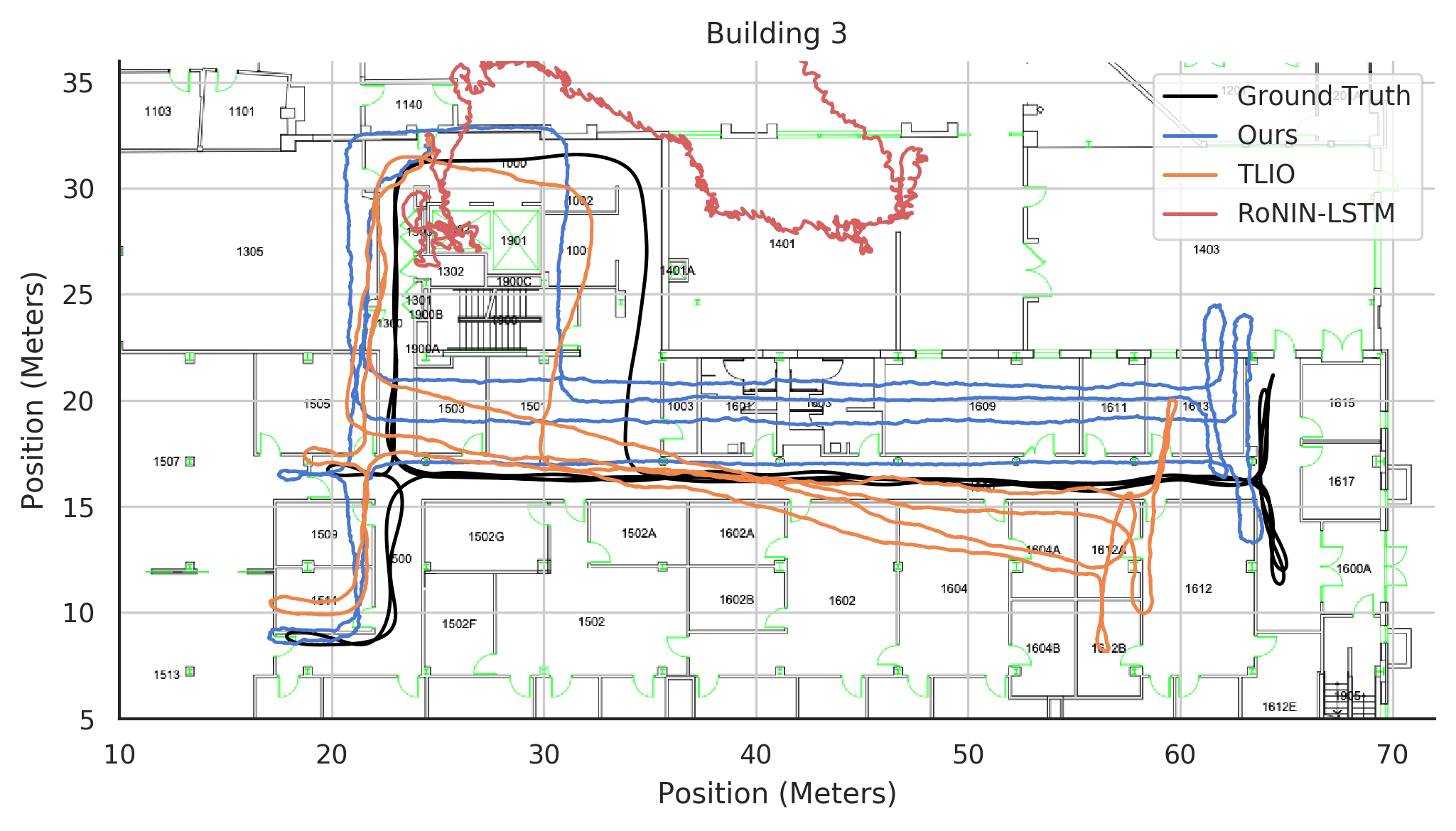}
    \caption[Predicted trajectories on building dataset]{Trajectory comparison among our's, TLIO, and RoNIN. Same initial pose and truncated slightly for clarity.}
    \label{fig:building1_traj}
\end{figure}

Figure \ref{fig:building1_traj} compares some example trajectories in all three buildings among our method, the best performing variant of RoNIN, and TLIO. This succinctly illustrates the importance of a good orientation estimator, as TLIO and RoNIN's use of the phone estimate results in rotational drift that compromises the resulting position estimate (to the extent that the trajectories can leave the floorplan entirely). 

Table \ref{tab:orient_net_vs_gt} examines the impact on position error from using the phone orientation, our orientation, and the ground truth orientation on TLIO and RoNIN. We can see that not only does our orientation module improve the performance of all other position models (quite significantly for RoNIN), but it also nearly reaches the theoretical maximum performance where ground truth orientations are directly provided. 

\subsection{Generalization}
In our experiments, we notice two distinct failure modes for model generalization to new environments. The first is orientation failure due to reliance on magnetic readings, which leads to degraded performance in environments with wildly different magnetic fields from training, e.g., in new buildings. The second is position failure due to variations in building shape/size. Buildings in our dataset vary in dimension from tens to hundreds of meters and in composition of sharp vs rounded turns. While the first mode affects our model due to reliance on magnetic data, we discovered that \textit{all} data-driven methods suffer from the second failure mode, regressing position inaccurately in cross-building evaluations (train on one location, test on held-out location). This is perhaps unsurprising, as data-driven methods are known to fail on out-of-distribution examples. Regarding magnetic field variations, as long as our model is trained on a building with magnetic distortions, it can produce accurate orientation estimates that far exceed existing methods. This is due to its ability to capture magnetic field variation, but comes at the cost of degraded generalization in novel environments.

\section{Conclusion} 
\label{sec:conclusion}
In this work, we present a novel data-driven model that is one of the first to integrate deep orientation estimation with deep inertial odometry. Our orientation module uses an LSTM and EKF to form a stable, accurate orientation estimate that outperforms traditional and data-driven techniques like CoreMotion, MUSE, and \citetalias{denoise_imu}. Combined with a position module, this end-to-end-system localizes better than previous methods across multiple buildings and users. In addition, our orientation module is a swap-in component capable of empowering existing systems with orientation performance comparable to visual-inertial systems in known environments. Lastly, we build a large dataset of device pose, spanning 20 hours of pedestrian motion across 3 buildings and 15 people. Existing traditional inertial odometry methods either use assumptions or constraints on the user's motion, while previous data-driven techniques use classical orientation estimates. A pertinent issue future work should address is generalization across buildings through further data collection in unique environments, data augmentation, or architectural modifications.

\bibliography{references}

\end{document}